\documentclass[10pt,twocolumn,letterpaper]{article}
\pdfoutput=1

\usepackage[pagenumbers]{cvpr} 

\usepackage{url}            
\usepackage{booktabs}       
\usepackage{amsfonts}       
\usepackage{nicefrac}       
\usepackage{microtype}      
\usepackage{color}
\PassOptionsToPackage{table}{xcolor}

\usepackage{graphicx}
\usepackage{multirow}
\usepackage{makecell}
\usepackage{listings}
\usepackage{bm}
\usepackage{amsmath}
\usepackage{amsfonts,amssymb}
\usepackage{wrapfig}
\usepackage{appendix}
\usepackage{algorithm}
\usepackage{algorithmic}
\usepackage{pifont}
\newcommand{\cmark}{\ding{51}}%
\newcommand{\xmark}{\ding{55}}%
\usepackage{pythonhighlight}
\usepackage{listings}
\definecolor{codegreen}{rgb}{0,0.6,0}
\definecolor{codegray}{rgb}{0.5,0.5,0.5}
\definecolor{codepurple}{rgb}{0.58,0,0.82}
\definecolor{backcolour}{rgb}{0.95,0.95,0.92}

\lstdefinestyle{mystyle}{
    language=Python,
    basicstyle=\small\ttfamily,
    keywordstyle=\color{blue},
    commentstyle=\color{codegreen},
    stringstyle=\color{red},
    breaklines=true,
    showstringspaces=false,
}
\usepackage{etoolbox}
\makeatletter
\AfterEndEnvironment{algorithm}{\let\@algcomment\relax}
\AtEndEnvironment{algorithm}{\kern2pt\hrule\relax\vskip3pt\@algcomment}
\let\@algcomment\relax
\newcommand\algcomment[1]{\def\@algcomment{\footnotesize#1}}
\makeatother

\definecolor{RoyalBlue}{RGB}{65, 105, 225}
\definecolor{DarkGreen}{RGB}{16, 136, 16}
\definecolor{DarkRed}{RGB}{225, 32, 32}
\definecolor{Yellow}{RGB}{230, 192, 109}
\definecolor{Black}{RGB}{84, 88, 94}

\definecolor{cvprblue}{rgb}{0.21,0.49,0.74}
\usepackage[pagebackref,breaklinks,colorlinks,allcolors=cvprblue]{hyperref}

\def\paperID{6310} 
\def\confName{CVPR}
\def\confYear{2025}

\title{Rethinking Lanes and Points in Complex Scenarios \\ for Monocular 3D Lane Detection}

\author{%
  Yifan Chang $^{1, 3}$, Junjie Huang $^{2}$, Xiaofeng Wang$^{1, 3}$, Yun Ye $^{2}$, \\{ Zhujin Liang $^{2}$, Yi Shan $^{2}$, Dalong Du $^{2}$, Xingang Wang $^{1, 4}$\thanks{Corresponding author.}} \\
  $^1$Institute of Automation, Chinese Academy of Sciences, $^2$PhiGent Robotics, $^3$UCAS\\
  $^4$Luoyang Institute for Robot and Intelligent Equipment, Luoyang, China\\
  \texttt{\{changyifan2023, wangxiaofeng2020, xingang.wang\}@ia.ac.cn} \\
  \texttt{junjie.huang@ieee.org} \\
}

\begin{document}
\maketitle
\begin{abstract}

  Monocular 3D lane detection is a fundamental task in autonomous driving. Although sparse-point methods lower computational load and maintain high accuracy in complex lane geometries, current methods fail to fully leverage the geometric structure of lanes in both lane geometry representations and model design. \textbf{In lane geometry representations}, we present a theoretical analysis alongside experimental validation to verify that current sparse lane representation methods contain inherent flaws, resulting in potential errors of up to 20 m, which raise significant safety concerns for driving. To address this issue, we propose a novel patching strategy to completely represent the full lane structure. To enable existing models to match this strategy, we introduce the \textbf{E}nd\textbf{P}oint head (EP-head), which adds a patching distance to endpoints. The EP-head enables the model to predict more complete lane representations even with fewer preset points, effectively addressing existing limitations and paving the way for models that are faster and require fewer parameters in the future. \textbf{In model design}, to enhance the model's perception of lane structures, we propose the \textbf{P}oint\textbf{L}ane attention (PL-attention), which incorporates prior geometric knowledge into the attention mechanism. Extensive experiments demonstrate the effectiveness of the proposed methods on various state-of-the-art models. For instance, in terms of the overall F1-score, our methods improve Persformer by \textbf{4.4} points, Anchor3DLane by \textbf{3.2} points, and LATR by \textbf{2.8} points. The code will be available soon.
\end{abstract}    
\section{Introduction}
  
  In recent years, advances in autonomous driving have significantly improved vehicle perception and control. However, accurately detecting 3D lanes remains a challenging task due to complex road geometries, occlusions, and the lack of sufficient 3D annotations~\cite{ma2024monocular}. This task is crucial for applications like trajectory planning~\cite{williams2022trajectory} and high-definition map construction~\cite{li2022hdmapnet, liu2023vectormapnet, qiao2023machmap, ding2023pivotnet}. However, detecting the geometric structure and spatial location of 3D lanes solely from monocular images is highly challenging due to the lack of depth information.  

  \begin{figure*}[t]
    \centering
    \includegraphics[width=1.0\linewidth]{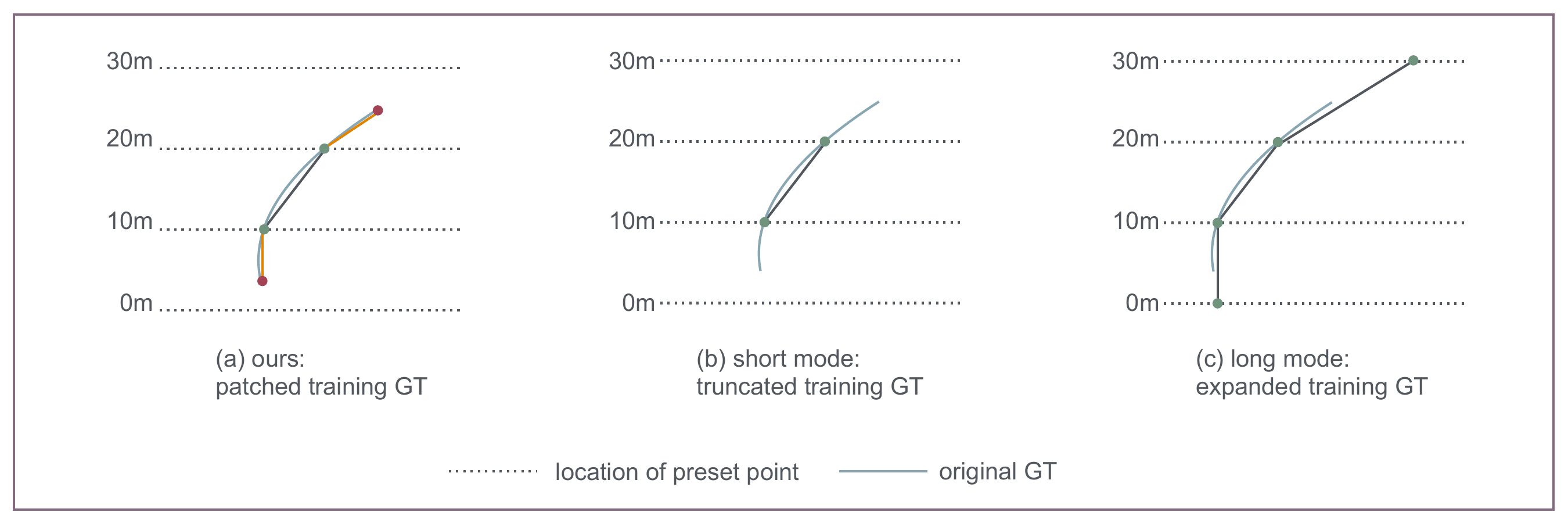}

   \caption{(a) Our method patches both endpoints and uses the EP-head to predict the patching distance, bringing the training ground truth closer to the original ground truth. (b) The short mode truncates both ends of the original ground truth, while (c) the long mode extends them. However, both methods produce inaccurate training ground truth and fail to fully capture the original ground truth.}
   \label{fig:intro}
  \end{figure*}

  To advance this task, various methods have been proposed, including pixel-wise~\cite{yan2022once}, grid-based~\cite{3dlanenet+, wang2023bevlanedet}, curve-based~\cite{bai2023curveformer, bai2024curveformer++, LaneCPP}, anchor-based~\cite{chen2022persformer,huang2023anchor3dlane}, and query-based approaches~\cite{luo2023latr}. Among these, anchor-based and query-based methods model lanes as sparse points, categorizing them as sparse-point methods. Sparse-point methods not only significantly lower computational load but also maintain high accuracy in complex lane geometries. However, existing sparse-point methods do not fully consider the geometric structure of lanes. In lane geometry representations, we find that existing training ground truth generation methods fail to represent the complete extent of the original ground truth. Specifically, during the generation of training ground truth, the original ground truth is truncated at the endpoints, as shown in \cref{fig:intro} (b). The truncated distance cannot be predicted by the model, which makes the model's results inherently inaccurate. In model design, these methods do not fully leverage prior geometric knowledge of lane structures. For example, Anchor3DLane~\cite{huang2023anchor3dlane} imposes parallel constraints on lanes, but these constraints are insufficient for handling real-world complex scenarios. Although LATR~\cite{luo2023latr} introduces 3D ground positional embedding priors and enhances cross-attention, it lacks more detailed modeling of lane structures and modifications to self-attention.

  To this end, we propose two modules to enhance existing methods. In lane geometry representations, we first introduce a novel endpoint patching strategy. This strategy increases the F1-score of training ground truth on the OpenLane dataset from 78.9\% to 98.5\%. To align existing models with this patching strategy, we propose the \textbf{EP-head}, which predicts the distances from each preset point to the endpoints of the original ground truth. After minimal processing, the EP-head yields more complete lane predictions. Its straightforward design allows for seamless integration into existing models, enhancing performance without substantial modifications. Additionally, the EP-head enables models to achieve more complete lane predictions even with fewer preset points, paving the way for future models that are faster and more parameter-efficient.

  Additionally, in model design, we propose \textbf{PL-attention} to better utilize the prior geometric knowledge of lanes. Lane detection tasks inherently possess prior geometric knowledge, which we summarize into three key aspects: a) On a single lane, adjacent points exhibit close relationships. Attributes (e.g., visibility) of neighboring points are highly likely to be similar, and the lane generally maintains overall smoothness. b) In a single image, relationships between lanes are closely linked. For instance, yellow-solid lines typically appear in the middle of the lane, while white-solid lines are usually located on the sides. c) Lanes at the same y-position share close relationships. Lanes on one road often have similar lengths and curvatures. PL-attention guides the model to learn features from three perspectives: within points on each lane, between different lanes, and among points at the same y-coordinate.
  
  The designed methods can be easily instantiated on existing methods for better performance. Moreover, extensive ablation studies have been conducted to elaborate on its superiority in effectiveness of each component, achieving consistent performance improvements with little computational overhead. For example, our approach improves the overall F1-score by 4.4 points for Persformer, 3.2 points for Anchor3DLane, and 2.8 points for LATR, thereby setting a new state-of-the-art performance.

Our primary contributions can be summarized as follows:

  1) We reveal, through theoretical analysis and experimental results, the flaws in current sparse lane representation methods, which have not been explored in previous research. Then, we propose a patching strategy using EP-head to patch endpoints, enabling the model to predict more complete lanes even with fewer preset points.

  2) We propose PL-attention which incorporates prior geometric knowledge related to points on each lane, interactions between different lanes, and connections among points at the same y-coordinate. This allows the model to utilize attention mechanisms to better perceive and adapt to diverse environments.

  3) Thorough experiments demonstrate the effectiveness of our methods. Our methods achieve an F1-score improvement of 4.4 points for Persformer, 3.2 points for Anchor3DLane, and 2.8 points for LATR, setting a new state-of-the-art performance.

\section{Related Work}

\subsection{Different Lane Geometry Representations}
How to model the lane geometry, specifically how to construct the training ground truth, is a crucial step in deep learning-based lane detection, which can be categorized into the following types: (1) Pixel-wise methods~\cite{neven2018seg1, zheng2021seg2, hou2019learning, qin2020ultra}, in 3D such as SALAD~\cite{yan2022once}, frame lane detection as a segmentation task by combining lane segmentation with depth prediction. These approaches, however, demand high computational resources due to the extensive number of parameters required. (2) Grid-based methods partition the space into cells, representing lanes with localized segments~\citet{huval2015empirical} or key points~\cite{qu2021key1, wang2022keypoint2, jin2022key3}. For instance, 3D-LaneNet+\cite{3dlanenet+} utilizes local line segments, while BEV-LaneDet\cite{wang2023bevlanedet} defines key points on a BEV (Bird’s Eye View) grid plane. Both methods rely heavily on grid resolution and involve costly post-processing to reconstruct continuous lanes. Together, pixel-wise and grid-based representations can be considered \textbf{dense} training ground truth representations since they require numerous parameters to describe the training ground truth. In contrast, the following methods represent \textbf{sparse} training ground truth representations: (3) Curve-based methods use predefined functions to model smooth lane curves, decoupling the position and shape of the lane~\cite{tabelini2021polylanenet, feng2022BezierLaneNet, van2019end, liu2020abcnet, bai2023curveformer, CLGo, 3d-splinenet, LaneCPP}. The lane’s position is determined by its start and end points (typically along the y-axis), while x-y and z-y functions capture its shape. However, relying solely on endpoints for positioning introduces a risk of misalignment: if the endpoint prediction is inaccurate, even a correctly predicted shape may lead to significant lane position errors. Notably, even though these approaches predict the lane endpoints, they still employ preprocessing steps in their code, as shown in \cref{fig:intro} (b), which truncates the original training ground truth. This highlights a gap in previous research, which has not prioritized maintaining the completeness of the training ground truth relative to the original training ground truth. (4) Sparse-point methods sample preset points within the detection range, a common approach in both anchor-based~\cite{zheng2022anchor1, liu2021rowanchor2, qin2022rowanchor3, 3d-lanenet, guo2020gen, huang2023anchor3dlane} and query-based models~\cite{laneformer, luo2023latr}. Nearly all existing models employ preprocessing (as shown in the top part of \cref{fig:intro}) to generate their training ground truth. As a result, they cannot capture the full extent of the original lane, and fewer preset points generally lead to increased error. Our approach (patching strategy with EP-head) address this gap, allowing current models to predict a complete lane even with fewer preset points.

\subsection{Geometry Priors}
Several methods have been proposed to incorporate prior knowledge into learning-based techniques, such as embedding invariance into the model architecture or utilizing task-specific transformations. In the lane detection task, for instance, LaneATT~\cite{laneatt} aggregates global features to infer location, while LATR~\cite{luo2023latr} predicts a 3D plane for positional embedding. SGNet~\cite{lu2021super} enforces constant lane width by penalizing lateral distance deviations in the IPM-transformed top-view, and LaneCPP~\cite{LaneCPP} learns parallelism and other geometric priors through analytical formulations of tangents and normals. Anchor3DLane~\cite{huang2023anchor3dlane} applies an equal-width constraint to adjust the x-coordinates of lane predictions. Simply considering lane parallelism and overall relationships cannot fully address the challenges posed by complex lane scenarios. We propose PL-attention, which optimizes the model structure by leveraging three types of relationships between lane points and lines, allowing the model to more effectively learn lane structures across a range of conditions.

\section{Preliminary Analysis}
\label{sec:Preliminary Analysis}
In this section, we first describe the issues we identify, then introduce the current mainstream evaluation methods. Next, we analyze in which scenarios the flaws in the lane representation method we identify are reflected in the performance metrics. Finally, we provide two commonly used datasets to validate our findings.

\subsection{Problem Description}
Lane datasets provide dense original ground truth, consisting of varying numbers of points (e.g., possibly fewer than 100 or more than 200). Such data cannot be directly used to train a model; instead, trainable ground truth (training GT) must be generated. To keep the model complexity low, we aim to reduce the number of points as much as possible. Generally, a rule is set to sample 10 or 20 preset points within the detection range. If a lane exists at a preset point, that point is marked as valid (visibility=1), and the x/z values at that point are retained. 
However, during the generation of training ground truth, the original ground truth is truncated twice at the endpoints, as shown in \cref{fig:intro} (b). The truncated distance can never be predicted by the model, making the model's results inherently inaccurate. The fewer the preset points representing the training ground truth, the greater the distance loss. If there are 20 preset points, up to 10 m may be lost. If there are 10 preset points, up to 20 m may be lost. Notably, this issue has not been explored before.

\subsection{Evaluation Method}
\label{sec:evaluation method}

During evaluation, we need to ensure that the ground truth and the predicted lanes have the same number of points. The more points we have, the more accurate our evaluation will be, with 100 preset points typically used. The original ground truth is first sampled at 100 preset points (3, 4, 5, 6, ..., 102, 103 m) to obtain the evaluating ground truth. Then, the predicted lane (training prediction) is sampled at the same 100 preset points to obtain the evaluating prediction. Finally, utilize bipartite matching to evaluate the evaluating ground truth and the evaluating prediction. If the evaluating ground truth's preset point (\eg the preset point at 5 m) and the evaluating prediction's corresponding point (at 5 m) are within 1.5 m, that point is considered matched. If 75\% of the preset points of a lane are matched, the matched lane is considered true positive. We refer to this 75\% threshold as Lane-IoU. Recall is determined by the percentage of matched ground-truth lanes, while precision is calculated based on the percentage of matched predicted lanes. By averaging both the recall and precision metrics, the F1-score is computed.~\cite{chen2022persformer}

\subsection{Theoretical Analysis}

Assume the training ground truth has 20 preset points, with an approximate distance of 5 m between each preset point. If one endpoint is truncated, the loss will be between 0 and 5 m; if both endpoints are truncated, the loss will be between 0 and 10 m. For simplicity of analysis, we assume the loss to be 10 m. If the lane length is less than 40 m, its training ground truth will be less than 30 m, and its training prediction will most likely be less than 30 m. When the Lane-IoU is 75\%, its evaluating prediction will be matched, but the evaluating ground truth will not, which means recall declines. We obtain the result from \cref{eq:analysis},

\begin{equation}
  \frac{x-10}{x} < 0.75, x < 40,
  \label{eq:analysis}
\end{equation}
where x denotes the length of lane.

We refer to this method of truncating both endpoints as the \textit{short mode}. In contrast, extending each endpoint to the next one is called the \textit{long mode}. \cref{fig:intro} (b) and (c) present the difference between the two modes. The effect of the \textit{long mode} is the opposite of the short mode--it causes the lane’s training ground truth to become longer, the training prediction to be longer, the evaluating ground truth to be matched, but the evaluating prediction not, which results in a decline in precision.

\begin{figure}[t]
    \centering
    \includegraphics[width=1.0\linewidth]{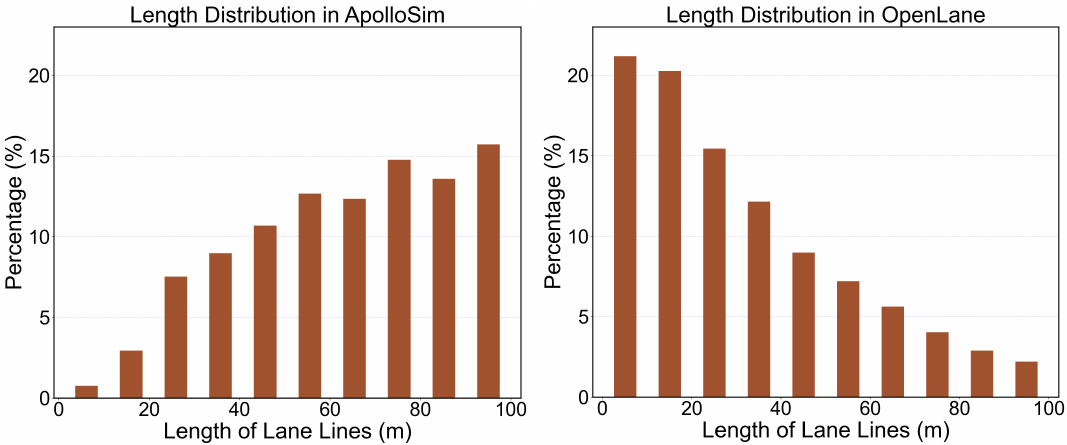}
    \caption{Lane length distribution comparisons between ApolloSim and OpenLane.}
    \label{fig:distribution}
\end{figure}

\begin{table}[t]
    \setlength{\tabcolsep}{1.9mm}
    \centering
    \caption{Results of training ground truth evaluated with different preset points (M).}
    \resizebox{\columnwidth}{!}{
    \begin{tabular}{c |c |c |c |c |c |c |c}
    \toprule[1pt]
        \multirow{2}{*}{M} & \multirow{2}{*}{mode} & \multicolumn{3}{c|}{OpenLane~\cite{chen2022persformer}} & \multicolumn{3}{c}{ApolloSim~\cite{guo2020gen}} \\
        \multirow{2}{*}{} & {}  & Rec $\uparrow$ & Pre $\uparrow$ & F1 $\uparrow$ & Rec $\uparrow$ & Pre $\uparrow$ & F1 $\uparrow$ \\
        \midrule[0.5pt]
        5 & long & 82.3 & 25.1 & 38.5 & 95.6 & 82.8 & 88.7 \\
        \midrule[0.5pt]
        \rowcolor{gray!30} 5 & short & 10.7 & 96.3 & 19.3 & 45.8 & 96.4 & 62.1 \\
        \midrule[0.5pt]
        10 & long & 94.2 & 48.7 & 64.2 & 98.3 & 91.0 & 94.5 \\
        \midrule[0.5pt]
        \rowcolor{gray!30} 10 & short & 35.4 & 98.8 & 52.1 & 84.4 & 97.3 & 90.4 \\
        \midrule[0.5pt]
        20 & long & 98.1 & 71.6 & 82.8 & 98.4 & 96.6 & 97.5 \\
        \midrule[0.5pt]
        \rowcolor{gray!30} 20 & short & 65.4 & 99.5 & 78.9 & 94.2 & 97.7 & 95.9 \\
        \midrule[0.5pt]
        40 & long & 99.3 & 86.6 & 92.5 & - & - & - \\
        \midrule[0.5pt]
        \rowcolor{gray!30} 40 & short & 82.2 & 99.7 & 90.1 & - & - & - \\
        \midrule[0.5pt]
        100 & long & 99.8 & 95.7 & 97.7 & - & - & - \\
        \midrule[0.5pt]
        \rowcolor{gray!30} 100 & short & 94.7 & 99.8 & 97.2 & - & - & - \\
        \bottomrule[1pt]
    \end{tabular}}
    \label{tab:motivation1}
\end{table}

\subsection{Experimental Validation}
\label{sec:Experimental Validation}
To validate our theoretical analysis, we evaluate the training ground truth on different datasets (ApolloSim\footnote{ApolloSim dataset has three different split settings. We only use Balanced Scene, but it is sufficient to illustrate the issue.}\cite{guo2020gen} and OpenLane\cite{chen2022persformer}). As shown in \cref{fig:distribution}, OpenLane contains a higher proportion of shorter lanes compared to ApolloSim. In OpenLane, lanes shorter than 20 m account for approximately 40\%, and those shorter than 40 m make up around 70\%. In contrast, in ApolloSim, lanes shorter than 40 m account for only 20\%. This significant difference will highlight the flaws in the current lane representation methods.

According to the evaluation setup in \cref{sec:evaluation method}, we evaluate the training ground truth generated by both the short mode and long mode methods. From \cref{tab:motivation1}, three phenomena can be observed: (1) fewer preset points result in a lower F1-score; (2) the long mode shows lower precision, while the short mode shows lower recall; and (3) comparing OpenLane to ApolloSim, the higher the proportion of short lanes, the lower the F1-score.
These phenomena all demonstrate that the training ground truth is inaccurate, and the current lane representation methods have significant flaws.

\section{Methods}
\label{sec:method}
In this section, we first describe our 3D lane representation method, which uses a patching strategy. And we then introduce EP-head, enabling the model to implement the patching strategy and address the issues outlined in \cref{sec:Preliminary Analysis}. At last, we present the detailed process of incorporating prior geometric knowledge with PL-attention.

\subsection{Lane Representation}
Given an input image $\mathbf{I}\in\mathbb{R}^{H\times W\times 3}$, the goal of Monocular 3D Lane Detection is to predict the position and category of lanes. A 3D lane is described by $M$ preset 3D points uniformly sampled along the y-coordinate, implying that there are $m-1$ equal intervals between the two endpoints. For each preset point in a single lane, there exists an attribute denoting its visibility status. Also, an attribute is assigned to describe the category of lanes. Mathematically, each 3D lane is defined as 
\begin{equation}
  \boldsymbol{l_i}=\{\boldsymbol{p_i}, c_i\},
  \label{eq:task}
\end{equation}
and
\begin{equation}
  \boldsymbol{p_i}=\{(x_i^j, y_i^j, z_i^j, \mathrm{vis}_i^j, )\}_{j=1}^M,
  \label{eq:format}
\end{equation}
where the first three elements represent the spatial coordinates of the point $\boldsymbol{p_i^j}$ in 3D space, $\mathrm{vis}_i^j$ represents the visibility of point $\boldsymbol{p_i^j}$, and $c_i$ represents the category of lane $\boldsymbol{l_i}$. If $\mathrm{vis}_i^j=1$, the preset point is considered valid.

With the patching strategy, each preset point during training requires a patching distance to both the start point and end point of the original ground truth. 
\begin{equation}
  \boldsymbol{p_i}=\{(x_i^j, y_i^j, z_i^j, \mathrm{vis}_i^j, \boldsymbol{s_i^j}, \boldsymbol{e_i^j})\}_{j=1}^M,
  \label{eq:patch format}
\end{equation}
\vspace{-0.45cm}
\begin{equation}
  \boldsymbol{s_i^j}=(s_{xi}^j, s_{yi}^j, s_{zi}^j),
  \label{eq:patch format s}
\end{equation}
\vspace{-0.45cm}
\begin{equation}
  \boldsymbol{e_i^j}=(e_{xi}^j, e_{yi}^j, e_{zi}^j).
  \label{eq:patch format e}
\end{equation}
During inference, we only need to add the patching distance to the first and last valid preset points.

\subsection{EP-head}
\label{sec:ep-head}

To mitigate the issue in \cref{sec:Preliminary Analysis}, we patch endpoints of training ground truth. Patching distances of each preset point have to be predicted because each preset point is possibly to become endpoint including start point and end point. We propose \textbf{E}nd\textbf{P}oint head (EP-head) to predict them. EP-head works in the same position as regression head and visibility head, which predict $\{\boldsymbol{x}_i, \boldsymbol{z}_i\}$ and $\boldsymbol{vis}_i$ of \textit{i-th} lane respectively. We assume that point-aware features of \textit{i-th} lane are denoted as $\mathbf{P}_i\in \mathbb{R}^{M\times C_p}$, where $M$ is the number of preset points on each training ground truth and $C_p$ is the channel number of point-aware features. Like other heads, EP-head can be implemented using MLP. EP-head inputs $\textbf{P}_i$ and outputs the distances of each preset point from two endpoints, which can be succinctly described using \cref{eq:head}. 
\begin{equation}
\label{eq:head}
\centering
\begin{split}
    &\mathbf{\hat{s}}_i, \mathbf{\hat{e}}_i = \mathrm{MLP}(\mathbf{P}_i), \\
\end{split}
\end{equation}
where $\mathbf{\hat{s}}_i$ and $\mathbf{\hat{e}}_i$ denote predicted distances from each preset point to start point and end point of original ground truth.

\textbf{Loss.} Its training loss is described in \cref{eq:lossep}. $\mathbf{s_i}$ and $\mathbf{e_i}$ are easily obtained when we process the data. Note we just describe the loss of \textit{i-th} lane.
\begin{equation}
\label{eq:lossep}
\centering
\begin{split}
    &\mathrm{Loss}_{ep} = \frac{1}{M}\sum (||\mathbf{\hat{s}_i} - \mathbf{{s}_i}||_1 + ||\mathbf{\hat{e}_i} - \mathbf{{e}_i}||_1),
\end{split}
\end{equation}

\begin{algorithm}
	\renewcommand{\algorithmicrequire}{\textbf{Input:}}
	\renewcommand{\algorithmicensure}{\textbf{Output:}}
	\caption{Inference of EP-head}
	\label{inferEP}
	\begin{algorithmic}[1]
        \REQUIRE $\mathbf{P}_i$ represents point-aware features of i-th lane, $\mathbf{y}_i$ represents y-axis preset points of i-th training ground truth, $\mathbf{x}_i$ and $\mathbf{z}_i$ represents predicted x and z distances, $\mathbf{\hat{vis}}_i$ represents predicted visibility of i-th prediction.
        \ENSURE  $\mathbf{y}^*_i$ 
		\STATE $\mathbf{\hat{s}}_i, \mathbf{\hat{e}}_i \leftarrow \mathrm{MLP}(\mathbf{P_i})$
		\STATE  $\mathbf{idx}_i \leftarrow$ find the indices of the first and last valid points from $\mathbf{\hat{vis}}_i$
            \STATE $\mathbf{x}^*_i, \mathbf{y}^*_i, \mathbf{z}^*_i   \leftarrow$ modify the $\mathbf{x}_i, \mathbf{y}_i, \mathbf{z}_i$ based on $\mathbf{idx}_i$, $\mathbf{\hat{s}}_i$ and $\mathbf{\hat{e}}_i$
	\end{algorithmic}  
\end{algorithm}

\textbf{Inference}. If the \textit{j-th} preset point of the \textit{i-th} lane is valid, namely $\mathrm{vis}_i^j=1$, and it is the first valid point, we add ($\hat{s}_{xi}^j$, $\hat{s}_{yi}^j$, $\hat{s}_{zi}^j)$ to ($x_i^j$, $y_i^j$, $z_i^j$). If the \textit{k-th} preset point of the \textit{i-th} lane is valid, namely $\mathrm{vis}_i^k=1$, and it is the last valid point, We add ($\hat{e}_{xi}^k$, $\hat{e}_{yi}^k$, $\hat{e}_{zi}^k$) to ($x_i^k$, $y_i^k$, $z_i^k$). For clarity, we describe it using code in Alg.~\ref{inferEP}. For simplicity, we do not consider the case where only one valid preset point exists. We only consider two or more valid preset points.

\subsection{PL-Attention}
\label{sec:pl-attention}
To help the model learn features more effectively, we introduce prior knowledge derived from our understanding of lane design. Lane detection differs from object detection~\cite{faster-rcnn, mask-rcnn, cornernet, detr, zhu2020deformable} in that it aims to represent the shape of lanes with a small number of points, rather than enclosing objects with bounding boxes. Lane detection tasks inherently possess some prior knowledge, which we summarize into three key aspects: within points on each lane, between different lanes, and among points at the same y-coordinate.

We propose \textbf{P}oint\textbf{L}ane attention (PL-attention) to introduce prior knowledge of lanes. PL-attention consists of three components, including point-point attention(PPA), lane-lane attention(LLA), and point-y attention(PYA). Specifically, we implement them using three simple multi-head self-attentions, but combining them in a way that conforms to the structure of lanes. Before PL-attention, we first reshape point-aware features $\mathbf{P}$ into ${N\times M\times C}$, where $N$ is the number of lanes, $M$ is the number of preset points and $C$ is channel number. Features of points on the \textit{i-th} lane are denoted as $[\mathbf{p}_{i,1}, \mathbf{p}_{i,2}, ..., \mathbf{p}_{i,M}]$. After the end of each lane, we insert a $\mathrm{[CLS]}$ token (denoted as $\mathbf{p}^s$), which is widely used in NLP~\cite{devlin2018bert, zhang2019hibert, wu2021hitransformer} and CV~\cite{dosovitskiy2020ViT, xu2022multicls} tasks. There, we use it to convey the contextual information within this lane.

\textbf{Point-point attention.} Position embeddings are added to features $\mathbf{P}$. features need to output X and Z positions, hence, position embeddings can not be preset. We use MLP to learn position embeddings described in \cref{eq:position}.
\begin{equation}
\label{eq:position}
\centering
\begin{split}
    &\mathbf{P}' = \mathbf{P} + \mathrm{MLP}(\mathbf{P}),
\end{split}
\end{equation}

\begin{figure}[t]
    \centering
    \includegraphics[width=1.0\linewidth]{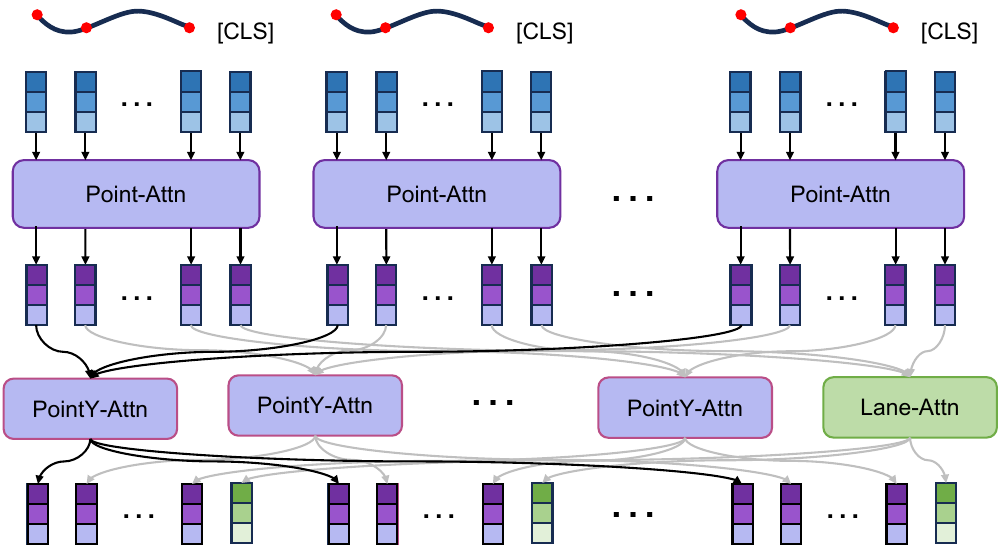}
    \caption{Overview flow of PL-attention with Point-point attention (Point-Attn), lane-lane attention (Lane-Attn) and point-y attention (PointY-Attn).}
    \label{fig:fusion}
\end{figure}

Point-point attention (PPA) takes $\mathbf{P}'_i$ as input to fully model the interactions between points on \textit{i-th} lane (\cref{eq:pp-attn}). Output are denoted as $\mathbf{P}^*_i=[\mathbf{p}^*_{i,1}, \mathbf{p}^*_{i,2}, ..., \mathbf{p}^*_{i,M}, \mathbf{p}^{s*}_{i}]$.
\begin{equation}
\label{eq:pp-attn}
\centering
\begin{split}
    &\mathbf{P}^*_i = \mathrm{PPA}(\mathbf{P}'_i),
\end{split}
\end{equation}

\textbf{Lane-lane attention.} Next, we extract lane-aware features as $\mathbf{P}^{s*}$, which entail the information of lanes, and we apply lane-lane attention (LLA) to them to capture the global information. The process is described in (\cref{eq:ll-attn}) and output is denoted as $\mathbf{H}^s=[\mathbf{h}^s_{1}, \mathbf{h}^s_{2}, ..., \mathbf{h}^s_{N}]$. 
\begin{equation}
\label{eq:ll-attn}
\centering
\begin{split}
    \mathbf{H}^s = \mathrm{LLA}(\mathbf{P}^{s*}),
\end{split}
\end{equation}
where $\mathbf{P}^{s*}\in \mathbb{R}^{N\times C}$, $\mathbf{H}^{s}\in \mathbb{R}^{N\times C}$. $\mathbf{P}^*$ without $\mathbf{P}^{s*}$ is denoted as $\mathbf{P}^{\dagger}$.

\textbf{Point-y attention.} To further provide prior knowledge to the model, we use Point-y attention (PYA) to flexibly constrain, rather than enforcing strict control as in ~\cite{huang2023anchor3dlane}. We reshape $\mathbf{P}^{\dagger}$ into $M\times N\times C$, so that points at the same y position are in the same group. Point-y attention takes $\mathbf{P}^{\dagger}_i$ as input and output $\mathbf{H}^{\dagger}_i$. This process is summarized in \cref{eq:py-attn}.
\begin{equation}
\label{eq:py-attn}
\centering
\begin{split}
    &\mathbf{H}^{\dagger}_i = \mathrm{PYA}(\mathbf{P}^{\dagger}_i),
\end{split}
\end{equation}

After PL-attention, lane-aware features $\mathbf{H}^s$ can be used to predict classification of lanes, instead of using max pooling in point-aware features as in ~\cite{luo2023latr}. 

\textbf{Complexity.} We implement PL-attention with three multi-head self-attentions~\cite{vaswani2017attention}. Its complexity is $O(NM^2+N^2+MN^2)$. In theory, its complexity is lower than sending all points into multi-head self-attention, whose complexity is $O(N^2M^2)$.

\textbf{How to apply it to existing methods.} Applying PL-attention to query-based methods is straightforward. For instance, simply replace the self-attention in LATR~\cite{luo2023latr} with PL-attention. For other methods, we put the PL-attention between abstracting lane feature and predicting lane attribute. The details will be discussed in the supplementary material.

\section{Experiments}
In this section, we first conduct the same experiment as in \cref{sec:Experimental Validation} to verify that our patching strategy produces a complete and accurate training ground truth. Next, we introduce the dataset and metrics, followed by implementation details. We then demonstrate the effectiveness of our method through comparisons with leading approaches. Finally, we present an analysis of each component.

\subsection{Effectiveness of Patching Strategy}
We evaluate the training ground truth after applying the patching strategy on OpenLane, and the results are shown in \cref{tab:patching result}. There is no abnormal decrease in recall and precision. Even with fewer preset points, a high F1-score is maintained. However, the F1-score still drops significantly when there are only 5 preset points. This occurs because, with 5 preset points, most training ground truth has only one valid preset point, and for simplicity, we currently consider only cases with two or more valid preset points, though it is straightforward to extend to a single preset point. These results indicate that the patching strategy can generate complete and accurate training ground truth.

\begin{table}[t]
    \centering
    \caption{Results of patched training ground truth evaluated with different preset points(M).}
    \resizebox{\columnwidth}{!}{
    \begin{tabular}{c |c |c |c |c |c |c |c }
    \toprule[1pt]
        \multirow{2}{*}{M} & \multirow{2}{*}{{Rec} $\uparrow$} & \multirow{2}{*}{{Pre} $\uparrow$} & \multirow{2}{*}{{F1} $\uparrow$} &\multicolumn{2}{c}{{X error (m)} $\downarrow$} &\multicolumn{2}{c}{{Z error (m)} $\downarrow$}\\
        \multirow{2}{*}{} &\multirow{2}{*}{} &\multirow{2}{*}{} &\multirow{2}{*}{} &  $near$ & $far$ & $near$ & $far$ \\
        \midrule[0.5pt]
        5 & 72.3 & 80.9 & 76.4 & 0.352 & 0.281 & 0.464 & 0.352\\
        \midrule[0.5pt]
        10 & 91.9 & 96.0 & 93.9  & 0.143 & 0.111 & 0.149 & 0.122 \\
        \midrule[0.5pt]
        20 & 97.7 & 99.1 & 98.5 & 0.053 & 0.043 & 0.051 & 0.051 \\
        \bottomrule[1pt]
    \end{tabular}}
    \label{tab:patching result}
\end{table}

\begin{table*}[tb]
    \setlength{\tabcolsep}{1.9mm}
    \centering
    \caption{Comparisons with other 3D lane detection methods on the OpenLane validation dataset. Note that the results are officially published.}
    \scalebox{1.0}{
    \begin{tabular}{c |c |c |c |c c |c c }
    \toprule[1pt]
        \multirow{2}{*}{Methods} & \multirow{2}{*}{Backbone} & \multirow{2}{*}{{F1} $\uparrow$} & \multirow{2}{*}{{Acc.} $\uparrow$} & \multicolumn{2}{c}{{X error (m)} $\downarrow$} &\multicolumn{2}{c}{{Z error (m)} $\downarrow$}  \\
        \multirow{2}{*}{} &\multirow{2}{*}{} &\multirow{2}{*}{} &\multirow{2}{*}{} & $near$ & $far$ & $near$ & $far$ \\
        \midrule[0.5pt]
        3DLaneNet~\cite{3d-lanenet} & VGG-16~\cite{simonyan2014vgg} & 44.1 & - & 0.593 & 0.494 & 0.140 & 0.195 \\
        GenLaneNet~\cite{guo2020gen} & ERFNet~\cite{romera2017erfnet} & 32.3 & - &  0.591 & 0.684 & 0.411 & 0.521 \\
        CurveFormer~\cite{bai2023curveformer} & EffNet-B7~\cite{tan2019efficientnet} & 50.5 & - & 0.340 & 0.772 & 0.207 & 0.651 \\
        BEV-LaneDet~\cite{wang2023bevlanedet} & ResNet-34~\cite{resnet} & 58.4 & - & 0.309 & 0.659 & 0.244 & 0.631 \\
        GroupLane~\cite{li2023grouplane} & ResNet-50~\cite{resnet} & 60.2 & 91.6 & 0.371 & 0.476 & 0.220 & 0.357 \\
        PVALane~\cite{zheng2024pvalane} & ResNet-50~\cite{resnet} & 62.7 & \textbf{93.4} & 0.232 & 0.259 & 0.092 & 0.118 \\
        \midrule[0.5pt]
        Persformer~\cite{chen2022persformer} & EffNet-B7~\cite{tan2019efficientnet} & 50.5 & 92.3 & 0.485 & 0.553 & 0.364 & 0.431 \\
        \rowcolor{gray!30}Persformer+ours & EffNet-B7~\cite{tan2019efficientnet} & 54.9 & 88.5 & 0.311 & 0.329 & 0.186 & 0.229 \\
        Anchor3DLane~\cite{huang2023anchor3dlane} & ResNet-18~\cite{resnet} & 53.1 & 90.0 & 0.300 & 0.311 & 0.103 & 0.139 \\
        \rowcolor{gray!30}Anchor3DLane+ours & ResNet-18~\cite{resnet} & 56.3 & 91.2 & 0.257 & 0.263 & 0.079 & 0.110 \\
        LATR~\cite{luo2023latr} & ResNet-50~\cite{resnet} & 61.9 & 92.0 & 0.219  & 0.259 & 0.075 & \textbf{0.104} \\
        \rowcolor{gray!30}LATR+ours & ResNet-50~\cite{resnet} & \textbf{64.7} & 92.8 & \textbf{0.205} & \textbf{0.255} & \textbf{0.074} & 0.105 \\
        \bottomrule[1pt]
    \end{tabular}}
    \label{tab:main-result-openlane}
\end{table*}

\begin{table*}[tb]
    \setlength{\tabcolsep}{1.9mm}
    \centering
    \caption{Comparisons with leading methods on OpenLane benchmark in different scenarios. The evaluation metric is F1-score.}
    \resizebox{\textwidth}{!}{
    \begin{tabular}{c |l |l |l |l |l |l |l }
    \toprule[1pt]
         \multirow{2}{*}{Methods} & \multirow{2}{*}{All} & Up \& & \multirow{2}{*}{Curve} & Extreme & \multirow{2}{*}{Night} & \multirow{2}{*}{Intersection} & Merge \&\\
         {} & {} & Down & {} & Weather & {} & {} & Split \\
        \midrule[0.5pt]
        Persformer~\cite{chen2022persformer} & 50.5 & 42.4 & 55.6 & 48.6 & 46.6 & 40.0 & 50.7 \\
        \rowcolor{gray!20}Persformer+ours &  \textbf{54.9} \small\textcolor{RoyalBlue}{$\uparrow$ 4.4} & \textbf{48.1} \small\textcolor{RoyalBlue}{$\uparrow$ 5.7} &  \textbf{58.0} \small\textcolor{RoyalBlue}{$\uparrow$ 3.6} & \textbf{55.6} \small\textcolor{RoyalBlue}{$\uparrow$ 7.0} &  \textbf{51.2} \small\textcolor{RoyalBlue}{$\uparrow$ 4.6} &  \textbf{44.4} \small\textcolor{RoyalBlue}{$\uparrow$ 4.4} & \textbf{55.0} \small\textcolor{RoyalBlue}{$\uparrow$\,4.3}\\
        Anchor3DLane~\cite{huang2023anchor3dlane} & 53.1 & 45.5 & 56.2 & 51.9 & 47.2 & 44.2 & 50.5 \\
        \rowcolor{gray!20}Anchor3DLane+ours &  \textbf{56.3} \small\textcolor{RoyalBlue}{$\uparrow$ 3.2} & \textbf{48.9} \small\textcolor{RoyalBlue}{$\uparrow$ 3.4} &  \textbf{59.2} \small\textcolor{RoyalBlue}{$\uparrow$ 3.0} & \textbf{54.7} \small\textcolor{RoyalBlue}{$\uparrow$ 2.8} &  \textbf{50.7} \small\textcolor{RoyalBlue}{$\uparrow$ 3.5} &  \textbf{47.9} \small\textcolor{RoyalBlue}{$\uparrow$ 3.7} & \textbf{54.4} \small\textcolor{RoyalBlue}{$\uparrow$\,3.9}\\
        LATR~\cite{luo2023latr} & 61.9 & 55.2 & 68.2 & 57.1  & 55.4 & 52.3 & 61.5 \\
        \rowcolor{gray!20}LATR+ours &  \textbf{64.7} \small\textcolor{RoyalBlue}{$\uparrow$ 2.8} & \textbf{55.3} \small\textcolor{RoyalBlue}{$\uparrow$ 0.1} &  \textbf{71.5} \small\textcolor{RoyalBlue}{$\uparrow$ 3.3} & \textbf{58.5} \small\textcolor{RoyalBlue}{$\uparrow$ 1.4} &  \textbf{58.5} \small\textcolor{RoyalBlue}{$\uparrow$ 3.1} &  \textbf{55.5} \small\textcolor{RoyalBlue}{$\uparrow$ 3.2} & \textbf{65.5} \small\textcolor{RoyalBlue}{$\uparrow$\,4.0}\\
        \bottomrule[1pt]
    \end{tabular}}
    \label{tab:main-result2}
    \vspace{-0.45cm}
\end{table*}

\subsection{Datasets and Metrics}
OpenLane~\cite{chen2022persformer} is a comprehensive benchmark for 3D lane detection, based on the Waymo Open dataset~\cite{sun2020Waymo}. It comprises 1000 segments with 200K frames, capturing diverse conditions at $1280\times 1920$ resolution. It includes complex lane structures, 5 weather types, and 14 categories. It is currently the most challenging lane detection dataset. Therefore, we select it to showcase our methods and perform ablation studies.

In assessing 3D lane detection, we follow OpenLane~\cite{chen2022persformer} metrics, framing it as a matching problem based on predictions and ground truth. In addition to the F1-score presented in \cref{sec:evaluation method}, local accuracy is evaluated using point-wise distance errors along the x-axis and z-axis, with lane points divided into near points and far points.

\subsection{Implementation Detail}
To verify the effectiveness of our proposed method, we conduct experiments on the LATR~\cite{luo2023latr} framework unless otherwise specified. The LATR provides a simple architecture that can be used to reveal the properties of our method. In the default configuration of LATR, it adopts max-pooling along the point dimension followed by MLP for lane category. For the lane category, We directly employ MLP on lane-aware features produced by PL-attention. For all our experiments, we utilize the AdamW~\cite{adamw} optimizer with a weight decay of 0.01. We set the learning rate to $2\times 10^{-4}$ and employ a cosine annealing scheduler. We utilize a total batch size of 64, distributed across 8 NVIDIA GeForce RTX 3090 GPUs. Training spans 24 epochs on OpenLane~\cite{chen2022persformer}.

\subsection{Application on State-of-the-art Method}
To demonstrate the superiority of our proposed method, we apply EP-head and PL-attention on several state-of-the-art methods. All the methods are trained with their original configuration. The weight of $loss_{ep}$ is set to 1.0. For lane category, MLP is used on lane-aware features produced by PL-attention. The experiment results are presented in \cref{tab:main-result-openlane}. Our method consistently outperforms the baseline across all metrics by a significant margin. For instance, when applying our method, LATR~\cite{luo2023latr} achieves a 2.8\% absolute gain in F over the baseline and a 0.8\% increase in accuracy, setting a new state-of-the-art. Additionally, we evaluate the performance in different scenarios. As shown in \cref{tab:main-result2}, our method achieves a better performance. Especially, our methods show particularly significant improvements in challenging scenarios, including curves (\eg LATR +3.3\%), night (+3.1\%), intersections (+3.2\%), and merge-split scenarios (+4.0\%). These improvements demonstrate that our methods substantially enhance the model's ability to identify lane structures in complex scenarios.

\subsection{Ablation Study}
\subsubsection{Module Ablations}
To demonstrate the impact of EP-head and PL-attention on the model's detection performance, we conduct ablation experiments on both two components. In \cref{tab:epandpla}, both EP-head (+1.2\%) and PL-attention (+0.6\%), even when used separately, can improve the model's performance. Particularly, when they are used together (+2.8\%), the results are even better, surpassing the individual improvements of both components.

Furthermore, we conduct additional experiments with the preset point count set to 10. These experiments demonstrate the impact of preset point count on model performance and the effectiveness of components when preset points are sparse. Results are shown in \cref{tab:epandpla}, with the effect being more pronounced when the preset point count is low (40.6\% \vs 54.8\%).

\Cref{tab:short} presents the evaluation results for short lanes (0-40m). It is evident that the EP-head improves recall by 3.2 percentage points (55.4\% \vs 58.6\%), indicating that the EP-head enables the model to predict more complete short lanes.

\begin{table}[t]
    \setlength{\tabcolsep}{1.9mm}
    \centering
    \caption{Ablation study on EP-head and PLA. (10) indicates LATR with 10 preset points. The evaluation metric is F1-score.}
    \resizebox{\columnwidth}{!}{
    \begin{tabular}{c |c |c |c |c }
    \toprule[1pt]
        EP-head & LATR & LATR(10) & LATR+PLA & LATR(10)+PLA\\
        \midrule[0.5pt]
        \xmark & 61.9 & 40.6 & 62.5 & 44.7 \\
        \cmark & 63.1 & 49.9 & 64.7 & 54.8 \\
        \bottomrule[1pt]
    \end{tabular}}
    \label{tab:epandpla}
\end{table}

\begin{table}[t]
    \centering
    \caption{Evaluated results of short lanes(0-40m) on OpenLane.}
    \scalebox{1.0}{
    \begin{tabular}{c |c |c |c }
    \toprule[1pt]
        Lane-IoU=0.75 & F1 & Rec & Pre \\
        \midrule[0.5pt]
        LATR & 43.2 & 55.4 & 35.4  \\
        \midrule[0.5pt]
        Latr+EP-head & 44.2 & 58.6 & 35.5 \\
        \midrule[0.5pt]
        Latr+EP-head+PLA & 45.4 & 58.9 & 36.8 \\
        \bottomrule[1pt]
    \end{tabular}}
    \label{tab:short}
\end{table}

\begin{table}[t]
    \centering
    \caption{Evaluated results of lanes on OpenLane with different Lane-IoU.}
    \scalebox{0.8}{
    \begin{tabular}{c |c |c  }
    \toprule[1pt]
        F1 & Lane-IoU=0.75 & Lane-IoU=0.90 \\
        \midrule[0.5pt]
        LATR & 61.9 & 43.6  \\
        \midrule[0.5pt]
        Latr+EP-head & 63.1 & 47.1 \\
        \midrule[0.5pt]
        Latr+EP-head+PLA & 64.7 & 48.1 \\
        \bottomrule[1pt]
    \end{tabular}}
    \label{tab:Lane-IoU}
\end{table}

As the Lane-IoU threshold increases, the impact of an incomplete training ground truth progressively extends to lanes of varying lengths, with the effect on longer lanes becoming increasingly evident in the metric. At a Lane-IoU threshold of 0.75, the metric primarily reflects the incompleteness in short lane predictions, while at a threshold of 0.90, it further highlights the incompleteness in longer lane predictions. With the use of the EP-head, we observe an improvement of 3.5 F1 points at Lane-IoU=0.9 and an improvement of 1.2 F1 points at Lane-IoU=0.75, as shown in \cref{tab:Lane-IoU}. This indicates that the EP-head has a positive effect on lanes of varying lengths.

\subsubsection{Effectiveness of EP-head}
EP-head can be seen as a new form of supervision, similar to point regression and classification, and we have experimented with this in \cref{tab:ep}. Just training with EP-head alone can improve model performance (+0.9\%). But patching is still a major improvement (+1.3\%), which also suggests that existing models are limited by training ground truth.

In \cref{tab:lossweigth}, further ablation experiments on the weight of $loss_{ep}$ show that setting it to 1.0 balances the multi-task loss on a similar scale, achieving the best performance.

\begin{table}[t]
    \centering
    \caption{Evaluated results with different weights of $loss_{ep}$.}
    \resizebox{\columnwidth}{!}{
    \begin{tabular}{c |c |c |c c |c c }
    \toprule[1pt]
        \multirow{2}{*}{weight}  & \multirow{2}{*}{{F1} $\uparrow$} & \multirow{2}{*}{{Acc.} $\uparrow$} & \multicolumn{2}{c}{{X error (m)} $\downarrow$} &\multicolumn{2}{c}{{Z error (m)} $\downarrow$}  \\
        \multirow{2}{*}{} &\multirow{2}{*}{} &\multirow{2}{*}{} & $near$ & $far$ & $near$ & $far$ \\
        \midrule[0.5pt]
        1.0 & 64.7 & 92.8 & 0.205 & 0.255 & 0.074 & 0.105 \\
        10 & 63.7 & 92.4 & 0.228 & 0.276 & 0.183 & 0.245 \\
        20 & 62.9 & 91.9 & 0.226 & 0.289 & 0.358 & 0.360 \\
        \bottomrule[1pt]
    \end{tabular}}
    \label{tab:lossweigth}
\end{table}

\begin{table}[t]
    \setlength{\tabcolsep}{1.9mm}
    \centering
    \caption{Results of EP-head with/without \textit{training} and \textit{patch}. \textit{Training} refers to using the EP-head during training, while \textit{Patching} denotes executing steps 2 and 3 of Alg.~\ref{inferEP} during inference.}
    \scalebox{1.0}{
    \begin{tabular}{c |c |c }
    \toprule[1pt]
        training & patching & F1 \\
        \midrule[0.5pt]
        \xmark & \xmark & 62.5 \\
        \cmark & \xmark & 63.4 \\
        \cmark & \cmark & 64.7 \\
        \bottomrule[1pt]
    \end{tabular}}
    \label{tab:ep}
\end{table}

\subsubsection{Effectiveness of PL-attention}
Different prior information brings varying degrees of information to the model, demonstrated in \cref{tab:pla}. When only points within the same lane are allowed to interact (PPA), the performance is poor, with only 55.3\% F1-score. However, with only lane-lane attention (LLA) or point-y attention (PYA), the performance is higher than with only PPA, demonstrating the superiority of lane-level prior interaction. Performance improves with more priors involved in guidance. When all three priors are simultaneously introduced, the F1-score reaches its highest (64.7\%).  

\begin{table}[t]
    \setlength{\tabcolsep}{1.9mm}
    \centering
    \caption{Results of different components for PLA.}
    \scalebox{1.0}{
    \begin{tabular}{c |c |c |c }
    \toprule[1pt]
         PPA & LLA & PYA & F1\\
        \midrule[0.5pt]
        \cmark & \cmark & \cmark & 64.7 \\
         \cmark & \cmark & {} & 63.9 \\
       \cmark & {} & \cmark & 63.5 \\
        {} & \cmark & \cmark & 61.9 \\
        \cmark & {} & {} & 55.3 \\
        {} & \cmark & {} & 57.2 \\
        {} & {} & \cmark & 61.5 \\
        \bottomrule[1pt]
    \end{tabular}}
    \label{tab:pla}
\end{table}

\section{Conclusion}
In this work, we identify flaws in current sparse lane representation methods, provide theoretical analysis, and support our findings through carefully designed experiments. To mitigate these issues, we propose a new sparse lane representation incorporating patching strategy and introduce the EP-head, which effectively integrates with existing models. Additionally, we apply PL-attention to embed lane prior geometric knowledge into the attention mechanism. Our experimental results confirm that these methods enhance the model's adaptability to complex scenarios. We suggest that our methods could potentially be applied to both 2D lane detection and high-definition map construction. Improved lane representation methods will also be explored in future work.
{
    \small
    \bibliographystyle{ieeenat_fullname}
    \bibliography{main}
}

\clearpage
\setcounter{page}{1}
\maketitlesupplementary

\section{Comparison of Different Training Ground Truth Representations in 3D Lane}
1) \textbf{Pixel-wise Representations}: Pixel-wise methods, such as SALAD~\cite{yan2022once}, frame lane detection as a segmentation task by combining lane segmentation with depth prediction. These approaches, however, demand high computational resources due to the extensive number of parameters required.

2) \textbf{Grid-based Representations}: Grid-based methods divide the space into cells and represent lanes using localized segments or key points. For instance, 3D-LaneNet+\cite{3dlanenet+} utilizes local line segments, while BEV-LaneDet\cite{wang2023bevlanedet} defines key points on a BEV (Bird’s Eye View) grid plane. Both methods rely heavily on grid resolution and involve costly post-processing to reconstruct continuous lanes. For example, BEV-LaneDet sets a grid size of 0.5x0.5 meters. 

\begin{figure}[t]
    \centering
    \includegraphics[width=0.8\linewidth]{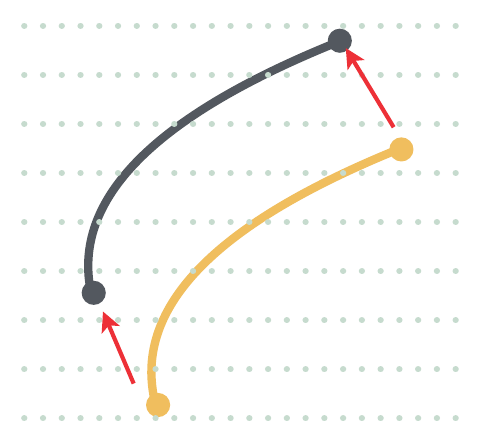}
    \caption{\textbf{Curve-based Representations}. \textcolor{Black}{The black lane} represents the training ground truth, while \textcolor{Yellow}{the yellow lane} represents the training prediction. Even if the shape of the prediction is accurate, any deviation in the endpoint position can cause the entire lane to shift significantly.}
    \label{fig:curve-flaws}
\end{figure}

Together, pixel-wise and grid-based representations can be considered \textbf{dense} training ground truth representations since they require numerous parameters to describe the training ground truth.
In contrast, the following methods represent \textbf{sparse} training ground truth representations:

3) \textbf{Curve-based Representations}: Curve-based methods use predefined functions to model smooth lane curves, decoupling the position and shape of the lane~\cite{bai2023curveformer, CLGo, 3d-splinenet, LaneCPP}. The lane’s position is determined by its start and end points (typically along the y-axis), while x-y and z-y functions capture its shape. However, relying solely on endpoints for positioning introduces a risk of misalignment: if the endpoint prediction is inaccurate, even a correctly predicted shape may lead to significant lane position errors. We illustrate this issue in \cref{fig:curve-flaws}. Notably, even though these approaches predict the lane endpoints, they still employ preprocessing steps in their code \footnote{\url{https://github.com/liuruijin17/CLGo/blob/main/db/apollosim_j.py}}, as shown in the top part of \cref{fig:intro}, which truncates the original training ground truth. This highlights a gap in previous research, which has not prioritized maintaining the completeness of the training ground truth relative to the original training ground truth.

4) \textbf{Sparse-point Representations}: Sparse-point methods sample preset points within the detection range, a common approach in both anchor-based~\cite{3d-lanenet, guo2020gen, huang2023anchor3dlane} and query-based models~\cite{luo2023latr}. Nearly all existing models employ preprocessing (as shown in the top part of \cref{fig:intro}) to generate their training ground truth. As a result, they cannot capture the full extent of the original lane, and fewer preset points generally lead to increased error.
Our approach (patching strategy + EP-head) address this gap, allowing current models to predict a complete lane even with fewer preset points. Our approach enables faster model inference (fewer preset points) while yielding more accurate training prediction. This work provides a novel perspective on lane representation, potentially guiding future research toward improved methodologies in this area.

  \begin{figure*}[t]
    \centering
    \includegraphics[width=1.0\linewidth]{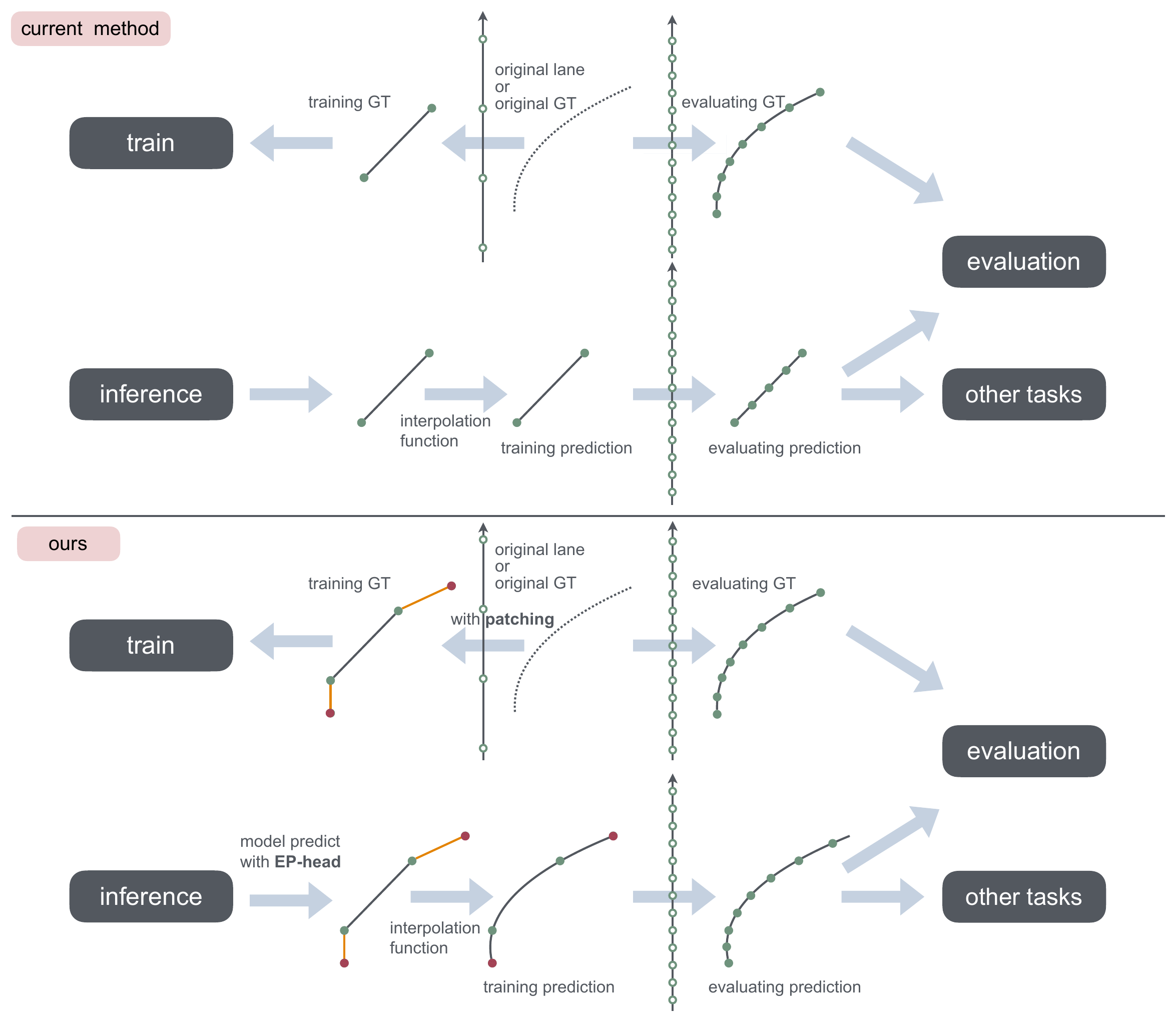}

   \caption{(1) The current sparse lane representation approach truncates the original ground truth, leading to an incomplete representation in the training ground truth. (2) In contrast, our method patches both endpoints and employs the EP-head to predict the patching distances, allowing the training prediction to more closely align with the original ground truth.}
   \label{fig:pipeline}
  \end{figure*}

\section{Pipeline of Lane Detection}
In \cref{fig:pipeline}, we present the pipeline of lane detection. 

\textbf{Training.} Lane datasets provide dense original ground truth, consisting of varying numbers of points (e.g., possibly fewer than 100 or more than 200). Such data cannot be directly used to train a model. We need to generate trainable ground truth (training ground truth). To keep the model complexity low, we aim to reduce the number of points as much as possible. Generally, a rule is set to sample preset points within the detection range (3-103m). Assuming there are 20 preset points, their positions will be at 3, 8, 13, ..., 98, and 103 meters in the forward direction. Specifically, the original ground truth is input into the interpolation functions, allowing for the calculation of the x/z values for any point within the lane. If a lane exists at a preset point, that point is marked as valid (visibility=1), and the x/z values at that point are retained. In this way, we obtain a new lane (called training ground truth) represented by 20 preset points. This lane can be effectively used to train the model. 

\textbf{Inference.} The format of the lane predicted by the model will be the same as the training ground truth. The prediction is then fed into the interpolation functions, and the output is referred to as the training prediction. The lane obtained by sampling additional points according to the task requirements is referred to as the evaluating prediction. 

\textbf{Evaluating.} During evaluation, we need to ensure that the ground truth and the predicted lanes have the same number of points. The more points we have, the more accurate our evaluation will be, with 100 preset points typically used. The original ground truth is first sampled at 100 preset points (3, 4, 5, 6, ..., 102, 103m) to obtain the evaluating ground truth. Then, the predicted lane (training prediction) is sampled at the same 100 preset points to obtain the evaluating prediction. Finally, utilize bipartite matching to evaluate the evaluating ground truth and the evaluating prediction. If the evaluating ground truth's preset point (\eg the preset point at 5m) and the evaluating prediction's corresponding point (at 5m) are within 1.5m, that point is considered matched. If 75\% of the preset points of a lane are matched, the matched lane is considered true positive. We refer to this 75\% threshold as Lane-IOU. Recall is determined by the percentage of matched ground-truth lanes, while precision is calculated based on the percentage of matched predicted lanes. By averaging both the recall and precision metrics, the F-score is computed.

\section{Implementation Details}
\label{appx:apply_pla}
\begin{figure*}[t]
    \centering
    \includegraphics[width=1.0\linewidth]{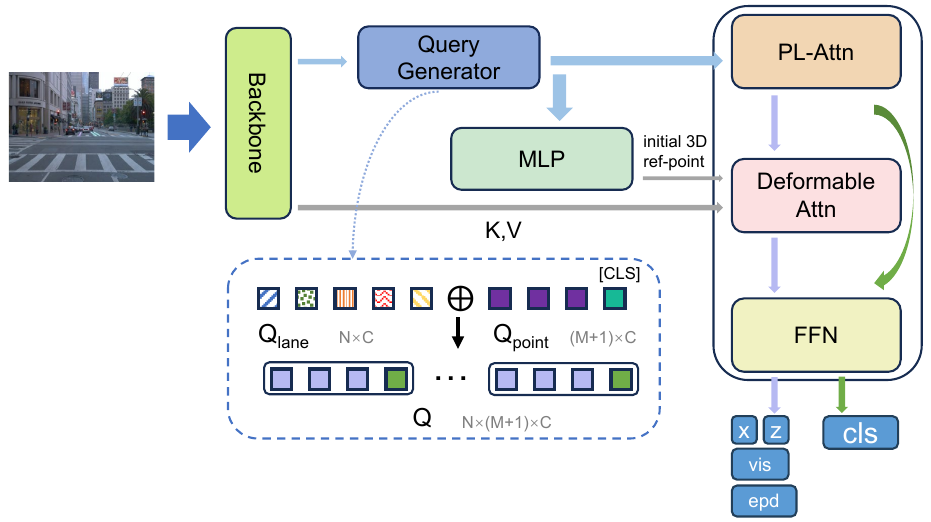}
    \caption{Overview of LATR with PL-attention and EP-head. The \textit{epd} means ep-distance that predictions from EP-head.}
    \label{fig:latr}
\end{figure*}

In \cref{fig:latr}, we provide an overview of LATR with PL-attention and EP-head. We make some modifications to LATR. Specifically, within the Query Generator, we insert a $[CLS]$ token into the point embedding $Q_{\text{point}}$. Subsequently, after the broadcasting summation, we obtain the lane-aware query embedding $Q\in \mathbb{R}^{N\times (M+1)\times C}$. Following the PL-attention step, the lane-aware features $\mathbf{H}^s$ are passed through a Feed-Forward Network (FFN), followed by a Multilayer Perceptron (MLP), to obtain the predicted classification ($cls$). Additionally, the point-aware features $\mathbf{P}^{\dagger}$ undergo Deformable Attention, followed by FFN and MLP, for predicted offsets ($x, z$), visibility ($vis$), and ep-distance ($epd$). Here, the ep-distance denotes the predictions from the EP-head.

Except for setting the weight of  $Loss_{ep}$  to 1.0, all other hyperparameters are kept identical to the original setup~\cite{luo2023latr}. Specifically, the total loss comprises three components: the instance segmentation auxiliary loss $L_{seg}$, the 3D ground plane update loss  $L_{plane}$, and the 3D lane prediction loss $L_{lane}$. Formally, these are represented as follows:

\begin{equation}
  L_{lane}=w_xL_x + w_zL_z + w_vL_v + w_cL_c + w_{ep}L_{ep},
\end{equation}

\begin{equation}
  L = w_sL_{seg} + w_pL_{plane} + w_lL_{lane},
\end{equation}

where  $w_{[*]}$  denotes the weights assigned to each loss component:  $w_s=5.0$, $w_x=2.0$, $w_z=10.0$, $w_c=10.0$, and all remaining weights are set to 1.0.  $L_x$  and  $L_z$  use L1 loss, while  $L_v$  employs binary cross-entropy (BCE) loss. For classification, focal loss~\cite{focalloss} is applied with parameters $\gamma =2.0$ and $\alpha =0.25$.

\section{Case for Patching Strategy and EP-head}
\begin{figure*}[t]
    \centering
    \includegraphics[width=1.0\linewidth]{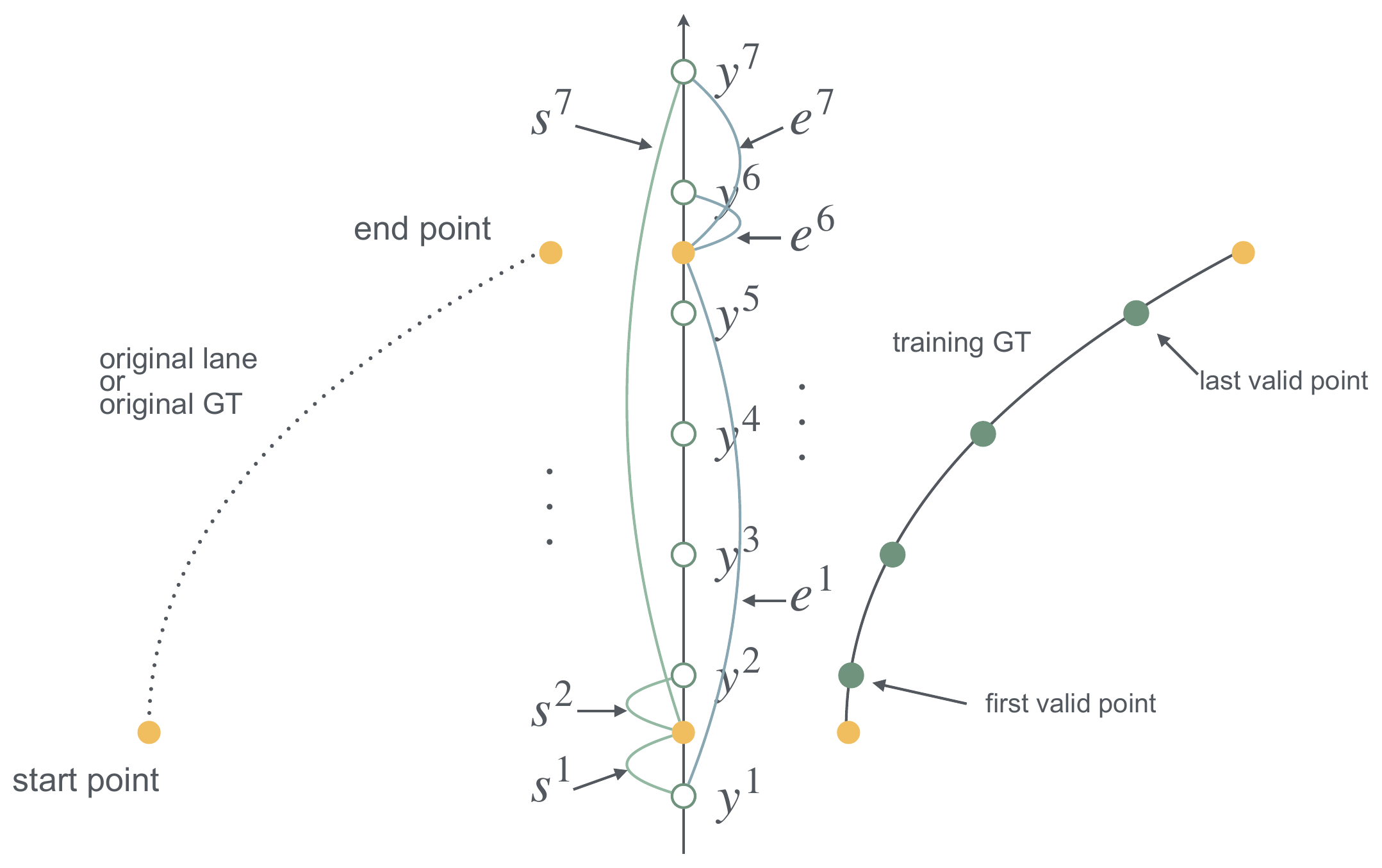}
    \caption{An illustrative case. The left lane represents the original ground truth, the middle indicates the detection direction, and the right lane is the training ground truth. Yellow dots mark the endpoints (start point and end point) of the original ground truth, green hollow dots denote preset points, and green solid dots indicate valid points.}
    \label{fig:case}
\end{figure*}

In \cref{fig:case}, an original ground truth consists of numerous points and is densely annotated. Resampling is performed based on preset points. After resampling, $y^2, y^3, y^4, y^5$ are valid points ($\mathrm{vis} = 1$), while $y^1, y^6, y^7$ are invisible points ($\mathrm{vis} = 0$). During inference, we do not know in advance which valid point will be the first or last valid point. Therefore, we need the model (via EP-head) to predict all distances $s_y^1, s_y^2, \dots, s_y^7$ and $e_y^1, e_y^2, \dots, e_y^7$. During post-processing, we identify the first and last valid points ($y^2$ and $y^5$) and update them by $y^{*2} = y^2 + s_y^2$ and $y^{*5} = y^5 + e_y^5$ (where $s_y^2$ is negative). This enables the model to predict a training prediction that closely resembles the original ground truth.

\section{More Ablation Study}

\subsection{PL-attention}
\begin{figure*}[t]
    \centering
    \includegraphics[scale=0.45]{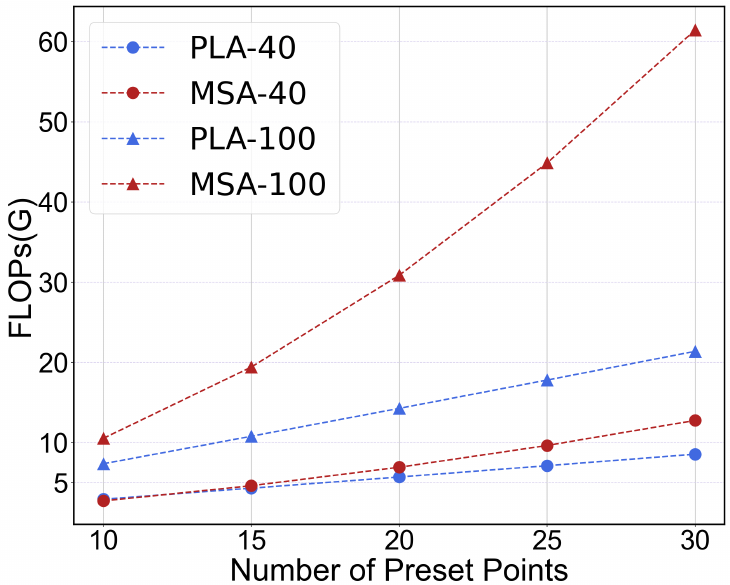}
    \caption{Complexity Comparison between PL-attention (PLA) and multi-head self-attention (MSA). \textit{PLA-40} refers to a preset number of lanes set to 40.}
    \label{fig:flops}
\end{figure*}

To demonstrate the complexity advantage of PLA, as mentioned in \cref{sec:pl-attention}, we analyze the complexity under different preset point counts and different preset numbers of lanes. \cref{fig:flops} illustrates that PL-attention (PLA) demonstrates lower computational complexity than multi-head self-attention (MSA) across various configurations of preset points. For instance, with 30 preset points, PLA-40 requires approximately 9G FLOPs, while MSA-40 approaches nearly 15G FLOPs. Similarly, PLA-100 utilizes around 20G FLOPs, in contrast to over 60G FLOPs for MSA-100. This trend indicates that the computational demand of PLA increases at a slower rate than that of MSA as the number of preset points grows, suggesting that PLA is more efficient.

\subsection{Visual Comparison}
\begin{figure*}[t]
    \centering
    \includegraphics[width=1.0\linewidth]{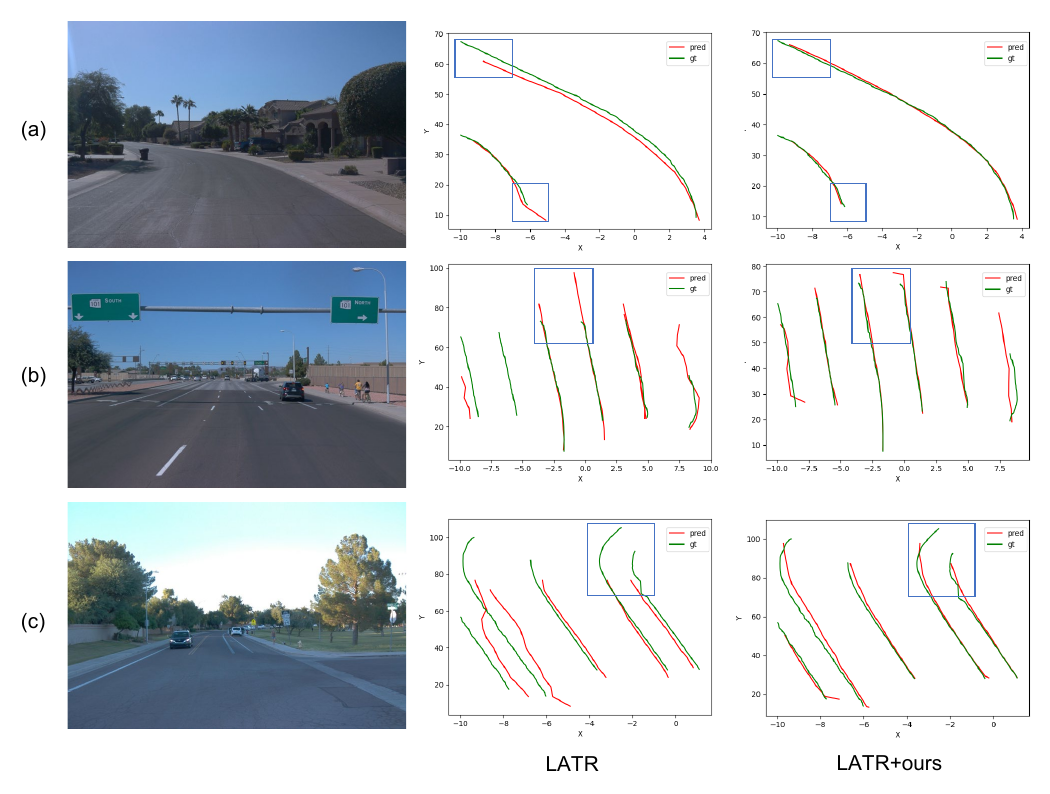}
    \caption{Qualitative evaluation on OpenLane val set. It presents the \textit{original ground truth} (\textcolor{DarkGreen}{green}) and the model's \textit{training prediction} (\textcolor{DarkRed}{red}) in 2D space.}
    \label{fig:vis}
    \vspace{-0.5cm}
\end{figure*}

Apart from quantitative results, we present qualitative examples in \cref{fig:vis}. In various scenarios, our method consistently predicts lanes with more complete endpoints and closer alignment to the original lane’s position and shape. (a) shows a single-lane turning scenario, where our predictions nearly perfectly match the original lane. (b) depicts a multi-lane, long-distance straight scenario, demonstrating our method's superior performance even with multiple lanes. (c) illustrates multi-lane turns, where LATR fails to capture the turn, while our method accurately predicts it.

\section{Discussion}

\begin{figure*}[t]
    \centering
    \includegraphics[width=1.0\linewidth]{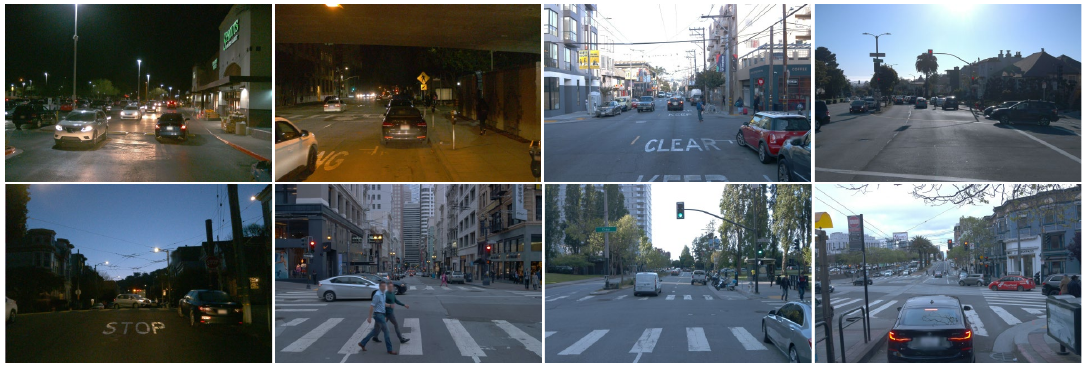}
    \caption{Scenarios with short lanes from OpenLane dataset.}
    \label{fig:scenarios}
\end{figure*}

\textbf{Complex scenarios with short lanes.} To observe scenarios where short lanes appear, we list some training data from OpenLane dataset in \cref{fig:scenarios}. The majority of scenes, including intersections, turns, and congested road sections, align with our experimental results..

\textbf{Why not directly set the ground truth on lanes.} Directly treating lanes as detection targets rather than presetting points would indeed result in ground truth accuracy reaching 100\%, regardless of whether the training ground truth is sparse or denser. However, this approach would introduce challenges during prediction. Specifically, predicting a correct sequence of points in a single image would become difficult. Additionally, if separating points by equal distance to address sequencing issues, errors in the y-direction could greatly amplify errors in the x/z-axis.

\textbf{Discussion on PL-attention.} We observe that in \cref{tab:pla} (1) standalone PPA yields the lowest results, yet PPA+PYA outperforms LLA+PYA; (2) LLA alone is weaker than PYA, but LLA+PPA surpasses PYA+PPA in performance. These inconsistencies primarily stem from the LATR's handling of point embeddings and lane embeddings. To better adapt LATR to PL-attention, we add a [cls] token to the point embeddings. After this, the original operations of LATR follow, where the augmented lane embeddings and point embeddings are added together. This results in the [cls] token carrying line embedding features, which represent the overall characteristics of lane during segmentation.

(1) The [cls] token also influences the point features through the lane embeddings, allowing LLA to incorporate some overall lane features and affect the point features, making LLA $>$ PPA. PPA+PYA is more focused on individual points, whereas LLA+PYA lacks modifications to the individual points of a lane itself.

(2) Both LLA+PPA (63.9) and PYA+PPA (63.5) possess overall lane features and individual point features, with a difference of only 0.4 between these two results, which is very small compared to other combinations. There are random variations here, making it difficult to analyze clearly.

\textbf{Limitation.} Patching methods also have limitations. Directly patching the endpoints can result in inaccurate predictions for the x/z direction at the endpoints. This influence can be seen in \cref{tab:main-result-openlane} and \cref{fig:vis} (b). The next phase of our research will focus on addressing inaccuracies in the x/z-axis. This research might serve as a breakthrough in predicting row lanes.

\section{Code}
\subsection{Code for Generating Training Ground Truth from Long Mode, Short Mode and Others}
In \cref{algo:lanerepresentation}, training ground truth generation code from 
Persformer~\cite{chen2022persformer}, Anchor3DLane~\cite{huang2023anchor3dlane}, and LATR~\cite{luo2023latr} can be viewed as a fix of both the long mode and short mode. They simply subtract or add 5 meters at both ends without considering the actual distance between the two points. However, the long mode and short mode have taken this into account. Additionally, we evaluate their training ground truth in \cref{tab:appendices}. The evaluation results are similar to those in \cref{tab:motivation1}.

\begin{table*}[ht]
    \setlength{\tabcolsep}{1.9mm}
    \centering
    \caption{Evaluated results of training ground truth from different methods.}
    \resizebox{\textwidth}{!}{
    \begin{tabular}{c |c |c |c |c |c |c |c |c }
    \toprule[1pt]
        \multirow{2}{*}{Method} & \multirow{2}{*}{M} & \multirow{2}{*}{{Rec} $\uparrow$} & \multirow{2}{*}{{Pre} $\uparrow$} & \multirow{2}{*}{{F1} $\uparrow$} &\multicolumn{2}{c}{{X error (m)} $\downarrow$} &\multicolumn{2}{c}{{Z error (m)} $\downarrow$}\\
        \multirow{2}{*}{} &\multirow{2}{*}{} &\multirow{2}{*}{} &\multirow{2}{*}{} &\multirow{2}{*}{} &  $near$ & $far$ & $near$ & $far$ \\
        \midrule[0.5pt]
        Persformer & 10 & 73.1 & 84.5 & 78.4  & 0.066 & 0.056 & 0.069 & 0.057 \\
        \midrule[0.5pt]
        Anchor3DLane\&LATR & 20 & 98.1 & 74.7 & 84.8 & 0.035 &  0.028 & 0.033 & 0.036 \\
        \bottomrule[1pt]
    \end{tabular}}
    \label{tab:appendices}
\end{table*}

\subsection{Code for Post-processing during Inference with Ep-head}
Implementing patching is straightforward and requires only a few additional lines of code in the post-processing step of the training predictions. We provide the modified code in \cref{algo:inferencepy}.

\onecolumn
\begin{algorithm}[t]
\caption{Pseudocode of generating training ground truth from long mode, short mode and others. }
\label{algo:lanerepresentation}
\definecolor{codeblue}{rgb}{0.25,0.5,0.5}
\lstset{
  backgroundcolor=\color{white},
  basicstyle=\fontsize{7.2pt}{7.2pt}\ttfamily\selectfont,
  columns=fullflexible,
  breaklines=true,
  captionpos=b,
  commentstyle=\fontsize{7.2pt}{7.2pt}\color{codeblue},
  keywordstyle=\fontsize{7.2pt}{7.2pt},
}
\begin{lstlisting}[language=python]
# Persformer
lane = [[x0, y0, z0], [x1, y1, z1], ..., [xn, yn, zn]]
anchor_y_steps = np.array([5, 10, 15, 20, 30, 40, 50, 60, 80, 100])
vis = np.logical_and(
    anchor_y_steps > lane[:, 1].min() - 5,
    anchor_y_steps < lane[:, 1].max() + 5)
# Anchor3DLane and LATR
num_y_steps = 20 #or 10
anchor_y_steps = np.linspace(3, 103, num_y_steps)
vis = np.logical_and(
    anchor_y_steps > lane[:, 1].min() - 5,
    anchor_y_steps < lane[:, 1].max() + 5)

# long mode
num_y_steps = 20 #or 10
anchor_y_steps = np.linspace(3, 103, num_y_steps)
vis = np.logical_and(
    self.anchor_y_steps >= lane[:,1].min() 
                    - 100.0/(num_y_steps-1),
    self.anchor_y_steps <= lane[:, 1].max() 
                     + 100.0/(num_y_steps-1))
# short mode
vis = np.logical_and(
    anchor_y_steps >= lane[:, 1].min(),
    anchor_y_steps <= lane[:, 1].max())

\end{lstlisting}
\end{algorithm}

\onecolumn

\begin{algorithm}[t]
\caption{Post-processing pseudocode during inference with ep-head.}
\label{algo:inferencepy}
\definecolor{codeblue}{rgb}{0.25,0.5,0.5}
\lstset{
  backgroundcolor=\color{white},
  basicstyle=\fontsize{7.2pt}{7.2pt}\ttfamily\selectfont,
  columns=fullflexible,
  breaklines=true,
  captionpos=b,
  commentstyle=\fontsize{7.2pt}{7.2pt}\color{codeblue},
  keywordstyle=\fontsize{7.2pt}{7.2pt},
}
\begin{lstlisting}[language=python]
# Post-processing code during inference with ep-head

xs, zs, vis, prob_c, deta_x_starts, deta_x_ends, deta_y_starts, deta_y_ends, deta_z_starts, deta_z_ends  = model(img, others)
num_y_steps = 20 #or 10
anchor_y_steps = np.linspace(3, 103, num_y_steps)
ys = np.tile(args.anchor_y_steps.copy()[None, :], (xs.shape[0], 1))

lanelines_pred = []
lanelines_prob = []

# xs.shape [number-preset-lane, number-preset-point]
for tmp_idx in range(xs.shape[0]):

    cur_vis = vis[tmp_idx] > 0
    cur_xs = xs[tmp_idx][cur_vis]
    cur_ys = ys[tmp_idx][cur_vis]
    cur_zs = zs[tmp_idx][cur_vis]
    # patching distance predicted by the model
    cur_deta_y_starts = deta_y_starts[tmp_idx][cur_vis]
    cur_deta_y_ends = deta_y_ends[tmp_idx][cur_vis]
    cur_deta_x_starts = deta_x_starts[tmp_idx][cur_vis]
    cur_deta_x_ends = deta_x_ends[tmp_idx][cur_vis]
    cur_deta_z_starts = deta_z_starts[tmp_idx][cur_vis]
    cur_deta_z_ends = deta_z_ends[tmp_idx][cur_vis]

    if cur_vis.sum() < 2:
        continue

    interval = 100.0 / (args.num_y_steps-1)

    # patching 
    cur_xs[0] = cur_xs[0] + cur_deta_x_starts[0]
    cur_xs[-1] = cur_xs[-1] + cur_deta_x_ends[-1]
    cur_ys[0] = cur_ys[0] + cur_deta_y_starts[0]
    cur_ys[-1] = cur_ys[-1] + cur_deta_y_ends[-1]
    cur_zs[0] = cur_zs[0] + cur_deta_z_starts[0]
    cur_zs[-1] = cur_zs[-1] + cur_deta_z_ends[-1]
    
    lanelines_pred.append([])
    for tmp_inner_idx in range(cur_xs.shape[0]):
        lanelines_pred[-1].append(
            [cur_xs[tmp_inner_idx],
             cur_ys[tmp_inner_idx],
             cur_zs[tmp_inner_idx]])
    lanelines_prob.append(prob_c[tmp_idx].tolist())
\end{lstlisting}
\end{algorithm}

\end{document}